\documentclass[journal]{IEEEtran}
\usepackage{color}
\usepackage{graphicx} 
\usepackage{amsmath}
\usepackage{amsfonts}
\usepackage{multirow}
\usepackage{makecell}
\usepackage{bbding}
\usepackage{lineno,hyperref}

\usepackage{xcolor}
\usepackage{soul}
\usepackage{array}    
\usepackage{colortbl}
\usepackage{algorithm}
\usepackage{algorithmic}

\usepackage[normalem]{ulem}
\useunder{\uline}{\ul}{}
\usepackage{booktabs}
\ifCLASSINFOpdf
\else
\fi
\usepackage{tikz,xcolor}
\hypersetup{hidelinks,
	colorlinks=true,
	allcolors=black,
	pdfstartview=Fit,
	breaklinks=true}
	
\definecolor{lime}{HTML}{A6CE39}
\DeclareRobustCommand{\orcidicon}{
\begin{tikzpicture}
\draw[lime, fill=lime] (0,0)
circle[radius=0.16]
node[white]{{\fontfamily{qag}\selectfont \tiny \.{I}D}};
\end{tikzpicture}
\hspace{-2mm}
}
\foreach \x in {A, ..., Z}{%
\expandafter\xdef\csname orcid\x\endcsname{\noexpand\href{https://orcid.org/\csname orcidauthor\x\endcsname}{\noexpand\orcidicon}}
}

\begin{document}
%
\title{
Decoupling Semantics and Fingerprints: A Universal Representation for AI-Generated Image Detection
}

%
%
%

\author{Zhiyuan~Wang\hspace{-1.5mm}\orcidA{},
        Yanxiang~Chen\hspace{-1.5mm}\orcidB{},~\IEEEmembership{Member,~IEEE},
        Pengcheng~Zhao\hspace{-1.5mm}\orcidC{},
        Yunfeng~Diao\hspace{-1.5mm}\orcidD{},
        Xin~Liao\hspace{-1.5mm}\orcidE{},~\IEEEmembership{Senior Member,~IEEE},
\thanks{Zhiyuan Wang is with Hefei University of Technology; 
(Email: zhiyuanwang@mail.hfut.edu.cn)
}
\thanks{Yanxiang Chen (Corresponding author) and Yunfeng Diao are with the Key Laboratory of Knowledge Engineering with Big Data (Hefei University of Technology), Ministry of Education; School of Computer Science and Information Engineering, Hefei University of Technology; and Intelligent Interconnected Systems Laboratory of Anhui Province (Hefei University of Technology).
(Email: chenyx@hfut.edu.cn; diaoyunfeng@hfut.edu.cn)
}
\thanks{Pengcheng Zhao is with the School of Computer Science, Nanjing Audit University.
(Email: 270783@nau.edu.cn)
}
\thanks{Xin Liao is with the College of Cyber Science and Technology, Hunan University, Changsha 410082, China.
(Email: xinliao@hnu.edu.cn)
}
}

%
%

\markboth{IEEE TRANSACTIONS ON INFORMATION FORENSICS AND SECURITY}%
{Wang \MakeLowercase{\textit{et al.}}: Decoupling Semantics and Fingerprints for AI-Generated Image Detection}
%



\maketitle
\begin{abstract}
Detecting AI-generated images across unseen architectures remains challenging, as existing models often overfit to generator-specific fingerprints and semantic content rather than learning universal forgery traces. We attribute this failure to feature entanglement: detectors learn these factors as a single entangled representation, where universal forgery traces are inextricably confounded with both generator-specific fingerprints and semantic content. Crucially, our spectral analysis reveals that this entanglement is avoidable: distinct generator-specific fingerprints (e.g., GAN stripes vs. Diffusion Model spots) occupy approximately disjoint frequency subspaces and coexist as independent superpositions. Leveraging this physical separability, we propose the Orthogonal Decomposition and Purification Network (ODP-Net) to structurally disentangle these factors.
Specifically, ODP-Net employs (1) Instance-aware Orthogonal Decomposition to project features into mutually exclusive subspaces: universal forgery traces, generator-specific fingerprints, and semantic content; (2) Counterfactual Purification to enforce semantic invariance via cross-sample feature injection; and (3) Manifold Alignment to bridge domain gaps.
By explicitly decoupling universal forgery traces from generator-specific fingerprints and semantic content, ODP-Net achieves state-of-the-art performance on unseen architectures (e.g., Stable Diffusion 3), validating that structural disentanglement is key to generalization.
\end{abstract}

\begin{IEEEkeywords}
Multimedia forensics, AI-Generated Image Detection, Feature Disentanglement, Orthogonal Decomposition, Cross-Generator Generalization.
\end{IEEEkeywords}

\IEEEpeerreviewmaketitle

\section{Introduction}

\begin{figure}[ht]
\centering
\includegraphics[width=3in]{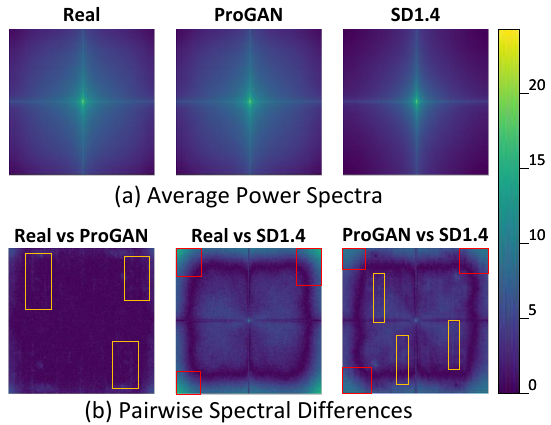}
\caption{Empirical evidence of spectral disparity and the physical basis for structural decomposition. (a) Average Power Spectra: Log-magnitude FFT spectra reveal that while Real, ProGAN, and SD1.4 share global statistics, they harbor distinct structural biases. (b) Pairwise Spectral Differences: Differential analysis uncovers unique generator-specific fingerprints. The orange rectangles highlight the vertical striping artifacts specific to ProGAN, while the red rectangles indicate the spectral spots characteristic of SD1.4. Notably, in the ProGAN vs. SD1.4 comparison, these distinct artifacts are superimposed, demonstrating that generator-specific fingerprints coexist as approximately disjoint patterns without interference. A rigorous quantitative validation of this spectral additivity is presented in Section~\ref{sec:spectral_justification}.}
\label{fig:fft_analysis}
\end{figure}

\IEEEPARstart{G}{enerative} Artificial Intelligence has seen significant advancements, enabling the synthesis of hyper-realistic images that blur the line between reality and forgery. 
While early GAN-based models \cite{goodfellow2014generative,karras2017progressive} and modern Diffusion Models \cite{ho2020denoising} have revolutionized content creation, they also pose severe risks to digital security. 
A primary challenge in forensic research is developing detectors that generalize to unseen generative architectures, as existing models often suffer from catastrophic performance degradation when encountering distribution shifts from novel generators \cite{cao2025survey}.

We argue that this lack of generalization stems from a fundamental structural misalignment: current detectors process forgery signals as a single, inextricably tangled web, failing to recognize that these signals are actually composed of independent, separable layers. To understand this, we must look at the frequency domain (as illustrated in our spectral analysis, \figurename{~\ref{fig:fft_analysis}}). Different generators leave distinct structural fingerprints—such as the vertical striping of ProGAN~\cite{karras2017progressive} or the high-frequency aliasing spots of Stable Diffusion~\cite{rombach2022high}. Intuitively, one might assume these various artifacts blend together into an inseparable noise. However, our differential analysis (detailed in Section~\ref{sec:spectral_justification}) reveals a structured phenomenon we term spectral additivity, where artifacts from different generators occupy largely disjoint frequency bands and approximately superimpose rather than destructively blending into unrecognizable noise. This property, which we quantify via the Additivity Criterion, confirms that distinct generators' spectral fingerprints occupy separable subspaces, providing the physical basis for the feature decomposition in Section~\ref{sec:method_decomposition}.

\begin{figure}[!t]
\centering
\includegraphics[width=2.8in]{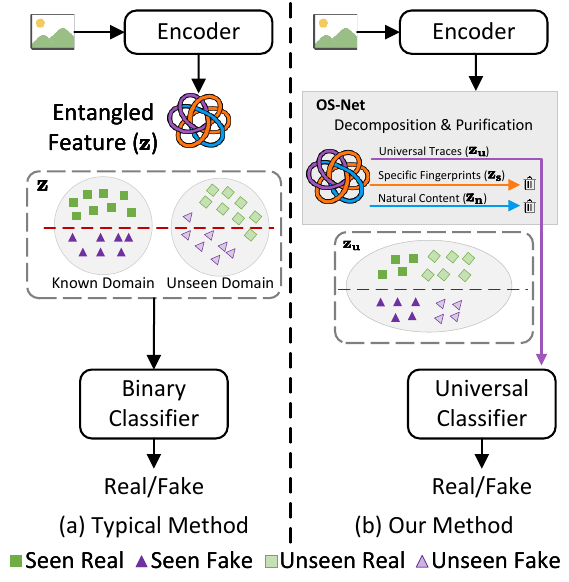}
\caption{Conceptual comparison of detection paradigms. \textbf{Left:} Typical methods extract entangled features where generator-specific fingerprints and semantic content dominate, resulting in overfitting to the source domain. \textbf{Right}: ODP-Net leverages the physical additivity of artifacts to explicitly disentangle universal traces from nuisance factors.}
\label{fig:comparison}
\end{figure}

To formally characterize these artifacts, we model the log-magnitude
Fourier spectrum. For a real image $\mathbf{x}_{\text{real}}$, the
spectrum follows the power-law statistics of natural scenes, denoted
by $\Phi$:
\begin{equation}
S_{\text{real}} \approx \Phi .
\end{equation}
An image from generator $k$ introduces an additive distortion term
$\Delta_k$---its ``fingerprint'':
\begin{equation}
S_{\text{gen}_k} \approx \Phi + \Delta_k .
\end{equation}
These fingerprints exhibit distinct topologies: ProGAN produces
periodic vertical patterns ($\Delta_{\text{stripe}}$) from upsampling
grids, while Stable Diffusion concentrates energy at specific frequency
extrema ($\Delta_{\text{spot}}$). We posit that these fingerprints are
spectrally disjoint: their frequency supports satisfy
$\text{supp}(\Delta_A) \cap \text{supp}(\Delta_B) \approx \emptyset$
for distinct architectures $A$ and $B$, implying the signals are
approximately orthogonal and exhibit additive superposition rather than destructive
interference.

Standard detectors ignore this physical reality. Instead, they learn an entangled representation where the crucial universal traces are inextricably mixed with specific generator fingerprints and the underlying semantic content of the image.
Consequently, as shown in \figurename{~\ref{fig:comparison}}, the decision boundary overfits to these spurious nuisance cues, leading to failure on unseen data where those specific fingerprints or semantics are absent. A rigorous quantitative validation of this spectral additivity is presented in Section~\ref{sec:spectral_justification}.

\begin{figure*}[!t]
\centering
\includegraphics[width=5.1in]{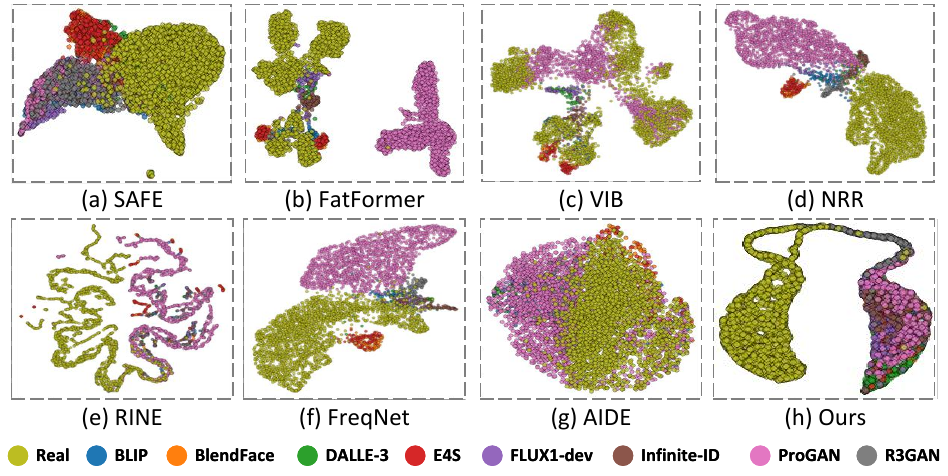}
\caption{UMAP visualization of the learned feature spaces on the training set. Yellow points denote Real images, while other colors represent various Fake sources (e.g., ProGAN, DALLE-3, Flux). Subfigures (a)-(g) reveal that baseline methods suffer from severe feature entanglement, where real and fake samples often overlap or fake samples fragment into generator-specific clusters. In contrast, (h) ODP-Net successfully aligns diverse forgeries into a compact, unified manifold that is strictly separable from the real domain, demonstrating robust generalization.}
\label{fig:umap_visualization}
\end{figure*}

The impact of this entanglement is evident in the latent space visualizations (\figurename{~\ref{fig:umap_visualization}}). 
Existing methods produce fragmented clusters (FatFormer\cite{liu2024forgery}) or dispersed distributions (AIDE\cite{yan2024sanity}), indicating that the model is classifying based on ``who generated the image" rather than ``whether the image is fake." To achieve true generalization, a detector must discard the ``who" and the ``what" (semantics) to focus solely on the ``whether" (universal traces).

To this end, we propose Orthogonal Decomposition and Purification Network (ODP-Net), a unified framework that bridges the gap between physical separability and feature representation. While spectral additivity is fundamentally a physical phenomenon observed in the frequency domain, this physical additivity does not automatically transfer to non-linear latent spaces. Therefore, we postulate that we must explicitly enforce this structural orthogonality within the high-dimensional feature representations. Driven by this insight, we design an objective that forces the latent space to be structurally decomposed.
ODP-Net achieves this through three synergistic stages.
Unlike previous methods that rely on implicit learning, we employ a hard masking mechanism that explicitly forces the feature vector to split into three orthogonal components: universal forgery traces ($\mathbf{z}_u$), generator-specific fingerprints ($\mathbf{z}_s$), and residual semantic nuisances ($\mathbf{z}_n$).

Second, to address the issue that decomposition alone might leave residual semantic correlations, we propose Counterfactual Purification.
By injecting cross-sample nuisance factors (a combination of $\mathbf{z}_s$ and $\mathbf{z}_n$ from donor images), we compel the network to unlearn semantic biases.
Finally, Manifold Alignment geometrically regularizes the purified features, contracting diverse generator clusters into a unified ``fake" prototype.

\begin{figure}[!t]
\centering
\includegraphics[width=2.8in]{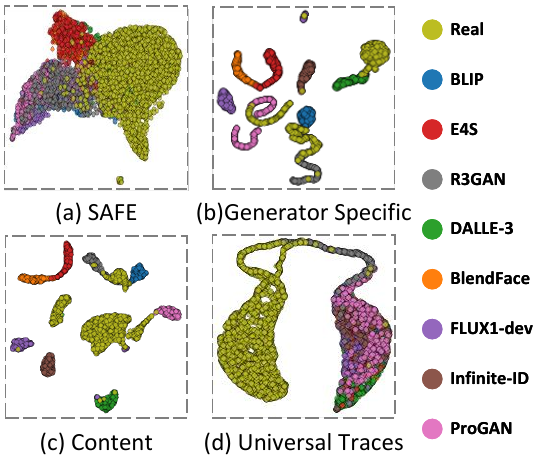}
\caption{UMAP visualizations of training-set feature distributions. Comparing (a) the state-of-the-art SAFE model with our disentangled components: (b) generator-specific fingerprints, (c) semantic content, and (d) universal forgery traces. While (b) and (c) display chaotic distributions, (d) demonstrates that our extracted universal traces form compact and strictly separated clusters, confirming effective disentanglement.}
\label{fig:feat_compare}
\end{figure}

The effectiveness of this structural refinement is substantiated in \figurename{~\ref{fig:feat_compare}}. 
While the separated generator-specific fingerprints (b) and semantic content (c) remain chaotic and entangled across domains, our extracted universal traces (d) converge into compact, well-separated clusters.
Supported by this disentanglement, ODP-Net maps diverse forgeries onto a single, unified manifold.

Our main contributions are summarized as follows:
\begin{itemize}
\item We discover and quantitatively validate the spectral additivity of generator artifacts, i.e., distinct forgery traces superimpose without interference in the frequency domain. This provides an empirical foundation for decoupling universal traces from generator-specific factors.
\item Guided by this physical prior, we propose ODP-Net, which first enforces orthogonal decomposition of latent features into disentangled subspaces, then refines the universal subspace via counterfactual purification and manifold alignment to suppress residual nuisance correlations that single-step separation cannot fully remove.
\item We demonstrate that ODP-Net significantly outperforms state-of-the-art methods in cross-generator generalization and probability calibration, confirming that physically grounded disentanglement, rather than implicit robustness, is the mechanism driving performance.
\end{itemize}

\section{Related Work}
\textbf{Generalized Forgery Detection.} Early detection methods primarily targeted semantic inconsistencies \cite{chen2022self,haliassos2021lips}. To improve generalization across generative models, research has shifted toward low-level signals presumed to be architecture-invariant, such as frequency statistics \cite{luo2021generalizing,qian2020thinking,tan2024frequency}, gradient patterns \cite{tan2023learning}, and diffusion reconstruction errors \cite{wang2023dire}. Despite these efforts, these methods often overfit to source-specific fingerprints (e.g., GAN grids), leading to performance degradation on unseen architectures. ODP-Net addresses this by structurally separating these specific fingerprints from universal forgery traces.

\textbf{Foundation Models in Forensics.}
Leveraging the semantic robustness of CLIP \cite{radford2021learning}, recent works have adapted Vision-Language Models for forensics. While early attempts used frozen features \cite{ojha2023towards}, subsequent approaches like FatFormer \cite{liu2024forgery} and C2P-CLIP \cite{tan2025c2p} introduced adapters to steer CLIP's focus toward high-frequency artifacts. However, these methods often fail to fully suppress CLIP's strong semantic bias, resulting in entangled representations where content interferes with detection. Our method resolves this by actively purifying semantic nuisance factors from the latent space.

\textbf{Feature Disentanglement.}
To mitigate feature entanglement, recent studies have explored subspace separation \cite{wang2025idcnet,wang2026precise}. VIB-Net \cite{zhang2025towards} utilizes information bottlenecks to filter redundancies, while Effort \cite{yan2024orthogonal} employs SVD to construct orthogonal subspaces for forgery features. Unlike these static or implicit decomposition methods, ODP-Net introduces instance-aware orthogonal decomposition combined with counterfactual purification. By dynamically masking features and enforcing structural orthogonality in the latent space, we explicitly disentangle universal traces from generator-specific and semantic factors, achieving superior generalization.

\begin{figure*}[!t]
\centering
\includegraphics[width=6in]{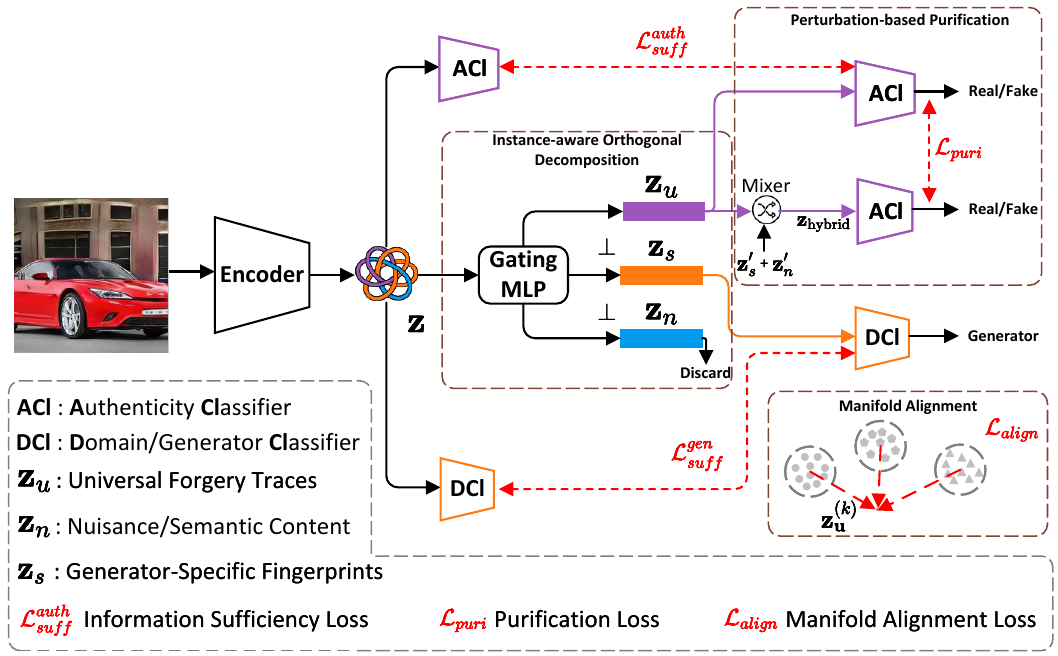}
\caption{
The pipeline consists of three stages: (1) Instance-aware Orthogonal Decomposition dynamically masks the latent feature $\mathbf{z}$ into three orthogonal components: universal traces ($\mathbf{z}_u$), specific fingerprints ($\mathbf{z}_s$), and residual nuisances ($\mathbf{z}_n$). (2) Counterfactual Purification enforces semantic invariance by injecting cross-sample nuisance features ($\mathbf{z}'_s + \mathbf{z}'_n$) from a donor into $\mathbf{z}_u$ and minimizing prediction inconsistency via $\mathcal{L}_{\text{puri}}$. (3) Manifold Alignment ($\mathcal{L}_{\text{align}}$) regularizes the geometry of the fake class clusters. ACl and DCl denote the Authenticity and Domain Classifiers, respectively.
}
\label{fig:framework}
\end{figure*}

\section{Methodology}

\subsection{Framework Overview}
Existing forgery detectors often learn entangled feature representations, coupling generator-specific fingerprints and semantic content with universal forgery traces. This entanglement hinders generalization to unseen generative models or diverse semantic contexts.

To explicitly address this, ODP-Net forces the latent feature vector $\mathbf{z}$ into a superposition of three mutually orthogonal components via structured hard masking:
\begin{equation}
\mathbf{z} = \mathbf{z}_u + \mathbf{z}_s + \mathbf{z}_n,
\label{eq:decomposition}
\end{equation}
where $\mathbf{z}_u \perp \mathbf{z}_s \perp \mathbf{z}_n$, with $\mathbf{z}_u$ encoding intrinsic, universal forgery traces invariant to synthesis methods and semantics; $\mathbf{z}_s$ capturing generator-specific fingerprints that uniquely characterize generative pipelines; and $\mathbf{z}_n$ corresponding to residual nuisance factors primarily arising from semantic variance.

The implementation of Eq.~\ref{eq:decomposition} is achieved through a three-stage pipeline. We first solve the structural separation problem using dynamic masking (Sec.~\ref{sec:method_decomposition}), followed by counterfactual purification (Sec.~\ref{sec:purification}), and conclude with a geometric regularization of the feature manifold (Sec.~\ref{sec:alignment}).

\subsection{Spectral Additivity as a Physical Prior for Decomposition}
\label{sec:spectral_justification}

Before detailing the decomposition mechanism, we first validate the physical basis that motivates it.
Visual inspection of spectral difference maps provides initial qualitative evidence. When we subtract the real spectrum from a GAN-generated image, we recover the isolated vertical striping pattern: $|S_{\text{real}} - S_{\text{gen}_{\text{GAN}}}| \approx |\Phi - (\Phi + \Delta_{\text{stripe}})| = |\Delta_{\text{stripe}}|$. More revealingly, the difference between two generators---e.g., ProGAN and Stable Diffusion---yields $|S_{\text{gen}_A} - S_{\text{gen}_B}| \approx |\Delta_{\text{stripe}} - \Delta_{\text{spot}}|$. As shown in \figurename{~\ref{fig:fft_analysis}}(b), this difference map simultaneously displays both the vertical stripes of ProGAN and the spectral spots of Stable Diffusion, rather than producing unrecognizable interference noise. This visual additivity suggests that distinct fingerprints occupy non-overlapping frequency supports.

To move beyond qualitative inspection, we quantitatively verify this separability using a rigorous linearity test. If two signals $\Delta_A$ and $\Delta_B$ are orthogonal (i.e., their spectral supports are disjoint), the magnitude of their difference must equal the sum of their individual magnitudes---the Additivity Criterion:
\begin{equation}
|\Delta_A - \Delta_B| \approx |\Delta_A| + |\Delta_B| .
\label{eq:additivity}
\end{equation}

We conduct a comprehensive verification on the AIGIBench \cite{li2025artificial} dataset. For every generator pair $(A,B)$, we compute three difference maps: the individual artifact maps $D_{A,\text{real}} = |S_{\text{gen}_A} - S_{\text{real}}|$ and $D_{B,\text{real}} = |S_{\text{gen}_B} - S_{\text{real}}|$, plus the cross-generator difference $D_{A,B} = |S_{\text{gen}_A} - S_{\text{gen}_B}|$. We then regress the predicted superposition $P_{\text{sum}} = D_{A,\text{real}} + D_{B,\text{real}}$ against the observed difference $D_{A,B}$. Orthogonal artifacts yield a strictly linear relationship with slope~1, quantified via the Pearson Correlation Coefficient (PCC) and Cosine Similarity (CS).

\begin{figure*}[t!]
\centering
\includegraphics[width=5.5in]{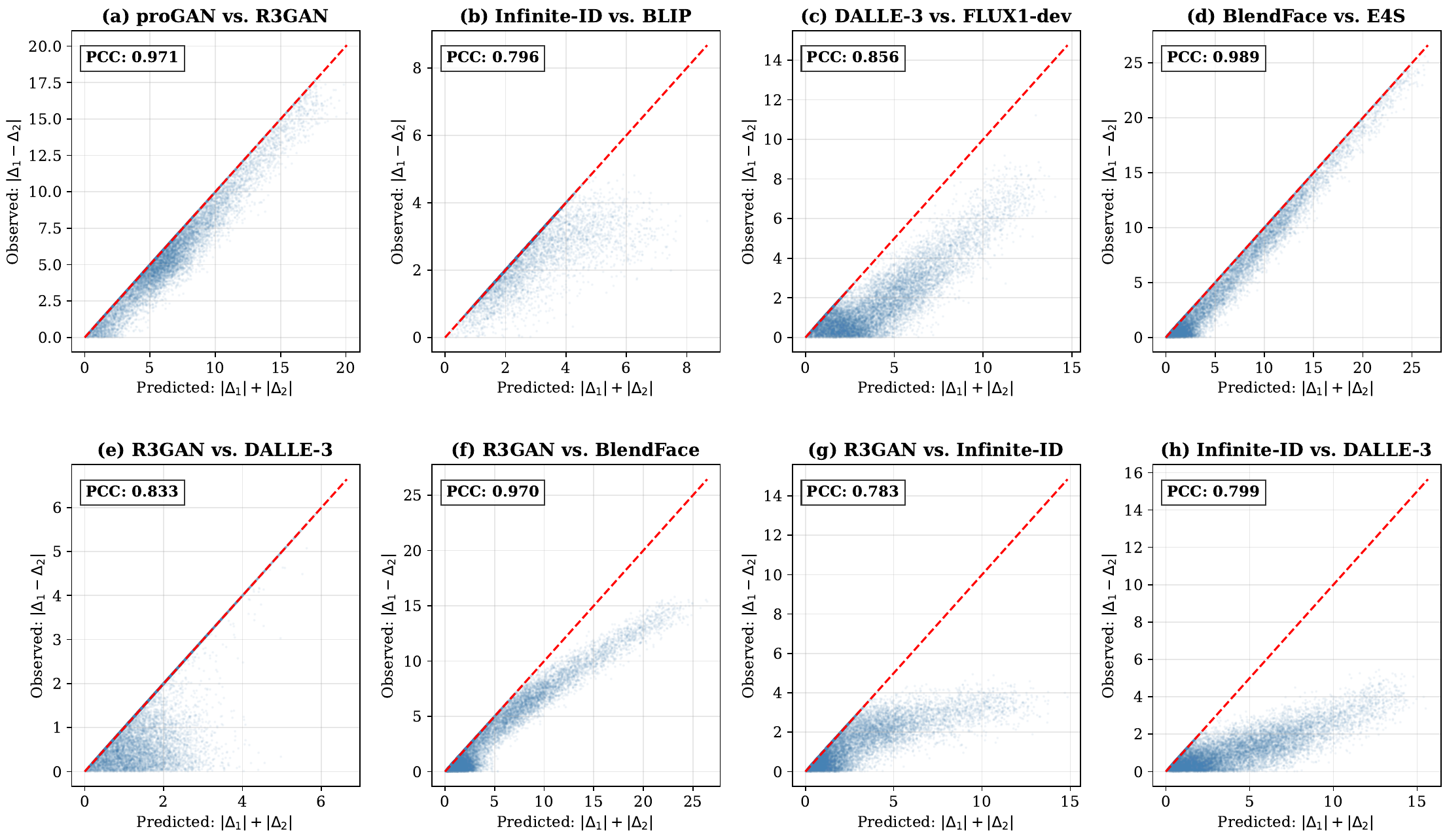}
\caption{Quantitative verification of spectral additivity via linearity of artifact superposition. We validate the hypothesis that artifacts from different generators occupy approximately disjoint spectral subspaces by plotting the predicted superposition magnitude ($|\Delta_1| + |\Delta_2|$) against the observed difference magnitude ($|\Delta_1 - \Delta_2|$). Ideally additive signals satisfy $|\Delta_1 - \Delta_2| \approx |\Delta_1| + |\Delta_2|$, corresponding to the diagonal red dashed line. The scatter plots (a-h) across diverse generator pairs (GANs, DMs, Deepfakes) exhibit strong linear correlation, empirically confirming that generator-specific fingerprints are approximately additive.}
\label{fig:linearity_verification}
\end{figure*}

\begin{table}[t!]
\centering
\caption{Quantitative validation of spectral additivity via the Additivity Criterion. We report the Pearson Correlation Coefficient (PCC) and Cosine Similarity (CS) between the sum of individual artifacts and the pairwise generator difference. High correlations confirm that artifacts from different sources are approximately additive, motivating the structural decomposition in ODP-Net.}
\label{tab:spectral_corr}
\resizebox{1.0\columnwidth}{!}{%
\begin{tabular}{@{}l|c|cc|c@{}}
\toprule
\multicolumn{1}{c|}{\multirow{2}{*}{\textbf{Generator Pair}}} & \multicolumn{1}{c|}{\textbf{Category Comparison}} & \multicolumn{2}{c|}{\textbf{Metrics}} & \multicolumn{1}{c}{\textbf{Hypothesis}} \\ \cmidrule(lr){2-5}
\multicolumn{1}{c|}{} & \multicolumn{1}{c|}{Model A vs.\ Model B} & \textbf{PCC} $\uparrow$ & \textbf{CS} $\uparrow$ & \textbf{Validation} \\ \midrule
\textbf{(a)} ProGAN vs.\ R3GAN & GAN vs.\ GAN & 0.971 & 0.992 & \cellcolor{green!15}Strong \\
\textbf{(b)} Infinite-ID vs.\ BLIP & DM-PG vs.\ DM-PG & 0.796 & 0.973 & Moderate \\
\textbf{(c)} DALLE-3 vs.\ FLUX1-dev & DM-T2IG vs.\ DM-T2IG & 0.856 & 0.934 & \cellcolor{green!15}Strong \\
\textbf{(d)} BlendFace vs.\ E4S & Deepfake vs.\ Deepfake & 0.989 & 0.992 & \cellcolor{green!15}Strong \\
\textbf{(e)} R3GAN vs.\ DALLE-3 & GAN vs.\ DM-T2IG & 0.833 & 0.926 & Moderate \\
\textbf{(f)} R3GAN vs.\ BlendFace & GAN vs.\ Deepfake & 0.970 & 0.986 & \cellcolor{green!15}Strong \\
\textbf{(g)} R3GAN vs.\ Infinite-ID & GAN vs.\ DM-PG & 0.783 & 0.911 & Moderate \\
\textbf{(h)} Infinite-ID vs.\ DALLE-3 & DM-PG vs.\ DM-T2IG & 0.799 & 0.922 & Moderate \\
\textbf{Avg.\ Correlation} & -- & \textbf{0.874} & \textbf{0.954} & \textbf{Validated} \\ \bottomrule
\end{tabular}
}
\end{table}

As shown in \figurename{~\ref{fig:linearity_verification}} and Table~\ref{tab:spectral_corr}, the additivity hypothesis is broadly validated with an average PCC of 0.874---though the correlation varies across generator pairs, ranging from strong (PCC 0.989 for BlendFace vs.\ E4S) to moderate (PCC 0.783 for R3GAN vs.\ Infinite-ID). This indicates that spectral additivity is an approximately true property rather than a universal law, with the strongest additivity observed among structurally distinct generators and some residual cross-talk within closely related diffusion models.

Nevertheless, the dominant additive relationship has an important consequence: it guarantees that artifacts can in principle be separated, i.e., an orthogonal basis for decomposing these signals exists in the signal space. However, this physical additivity does not automatically transfer to non-linear latent spaces such as that of a CLIP vision transformer. In fact, standard detectors severely entangle these factors because they lack the structural inductive bias to separate them. Therefore, we cannot rely on the feature space inheriting this separability---we must explicitly force it via a learnable hard masking mechanism. The Instance-aware Orthogonal Decomposition module below is precisely this forced projection: it uses the spectral additivity as a feasibility prior (the separation is possible) while the hard masking operation makes it mandatory. Extended analysis is provided in Appendix~\ref{appendix:spectral_verification}.

\subsection{Instance-aware Orthogonal Decomposition}
\label{sec:method_decomposition}

We construct a lightweight gating network that estimates channel-wise selection probabilities $\mathbf{P}_u, \mathbf{P}_s \in [0,1]^D$, representing the importance of each dimension to the universal and specific subspaces, respectively. To derive discrete masking patterns $\mathbf{M}_u, \mathbf{M}_s \in \{0,1\}^D$ amenable to backpropagation, we utilize the Straight-Through Estimator (STE) \cite{liu2022nonuniform}:
\begin{equation}
\mathbf{M}_k = \mathbf{P}_k + \mathrm{sg}\big(\mathbb{1}\{\mathbf{P}_k > 0.5\} - \mathbf{P}_k\big), \quad k \in \{u,s\},
\label{eq:ste_mask}
\end{equation}
where $\mathbb{1}\{\cdot\}$ denotes the indicator function and $\mathrm{sg}[\cdot]$ stops gradients.

Enforcing mutual exclusivity and structural orthogonality, we extract the three components via a cascaded masking operation:
\begin{equation}
\begin{aligned}
\mathbf{z}_u &= \mathbf{z} \odot \mathbf{M}_u, \\
\mathbf{z}_s &= \mathbf{z} \odot (1 - \mathbf{M}_u) \odot \mathbf{M}_s, \\
\mathbf{z}_n &= \mathbf{z} \odot (1 - \mathbf{M}_u) \odot (1 - \mathbf{M}_s),
\end{aligned}
\label{eq:cascaded_decomposition}
\end{equation}
with $\odot$ denoting element-wise multiplication.

To discourage degenerate solutions where masks trivially select all or no features, we impose a sparsity regularization:
\begin{equation}
\mathcal{L}_{\text{sparse}} = \frac{1}{D} \left( \|\mathbf{P}_u\|_1 + \|\mathbf{P}_s\|_1 \right),
\label{eq:sparsity}
\end{equation}
which encourages concise and meaningful channel selections.

Moreover, to guarantee that $\mathbf{z}_u$ and $\mathbf{z}_s$ retain sufficient information for forgery detection and generator identification, respectively, we adopt a self-distillation objective \cite{UrpBagVlaMar24}. Specifically, the authenticity classifier $f_u$ and generator classifier $f_s$ are trained on the decomposed features, with predictions guided to align with those from the undecomposed full feature $\mathbf{z}$. This is formalized as:
\begin{equation}
\begin{aligned}
\mathcal{L}_{\text{suff}}^{auth} &= D_{\mathrm{KL}}\big(f_u(\mathbf{z}_u) \parallel \mathrm{sg}[f_u(\mathbf{z})]\big), \\
\mathcal{L}_{\text{suff}}^{gen} &= D_{\mathrm{KL}}\big(f_s(\mathbf{z}_s) \parallel \mathrm{sg}[f_s(\mathbf{z})]\big),
\end{aligned}
\label{eq:sufficiency}
\end{equation}
where $\mathrm{sg}[\cdot]$ detaches gradients to treat full-feature predictions as fixed soft targets.

The decomposition module structurally disentangles universal traces from generator-specific fingerprints and semantic content via dynamic hard masking.

\subsection{Counterfactual Purification}
\label{sec:purification}

While the decomposition stage achieves structural separation, the extracted $\mathbf{z}_u$ may still harbor spurious semantic correlations. 
To explicitly enforce semantic invariance, we propose a counterfactual purification mechanism. 
For each sample with feature $\mathbf{z}$, we sample a donor feature $\mathbf{z}'$ from the minibatch (ensuring $\mathbf{z}' \neq \mathbf{z}$) and extract its nuisance component:
\begin{equation}
\mathbf{z}_{nuis}' = \mathbf{z}' \odot (1 - \mathbf{M}_u'),
\label{eq:donor_nuisance}
\end{equation}
where $\mathbf{M}_u'$ is the universal mask for $\mathbf{z}'$. We then synthesize a hybrid feature by combining the target's universal trace $\mathbf{z}_u$ with the donor's nuisance:
\begin{equation}
\mathbf{z}_{hybrid} = \mathbf{z}_u + \mathbf{z}_{nuis}'.
\label{eq:hybrid_feature}
\end{equation}

This hybrid simulates a composite instance sharing intrinsic forgery traces but possessing altered semantic and fingerprint content. A purification loss enforces that the authenticity classifier yields consistent predictions on $\mathbf{z}_u$ and $\mathbf{z}_{hybrid}$:
\begin{equation}
\mathcal{L}_{\text{puri}} = \| f_u(\mathbf{z}_u) - f_u(\mathbf{z}_{hybrid}) \|_2^2.
\label{eq:purification_loss}
\end{equation}

Minimizing $\mathcal{L}_{\text{puri}}$ penalizes reliance on nuisance-correlated channels. To understand how $\mathcal{L}_{\text{puri}}$ achieves this invariance, consider a first-order Taylor expansion of $f_u(\mathbf{z}_u + \boldsymbol{\delta})$ around $\mathbf{z}_u$, where $\boldsymbol{\delta} = \mathbf{z}_{nuis}'$ is the perturbation direction:
\begin{equation}
f_u(\mathbf{z}_u + \boldsymbol{\delta}) = f_u(\mathbf{z}_u) +
\nabla_{\mathbf{z}} f_u(\mathbf{z}_u)^\top \boldsymbol{\delta} +
\mathcal{O}(\|\boldsymbol{\delta}\|^2),
\end{equation}
Substituting into Eq.~\ref{eq:purification_loss} and neglecting higher-order terms yields:
\begin{equation}
\mathcal{L}_{\text{puri}} \approx \frac{1}{2}
\|\nabla_{\mathbf{z}} f_u(\mathbf{z}_u)^\top \boldsymbol{\delta}\|_2^2 .
\end{equation}
Minimizing this objective drives the dot product
$\nabla_{\mathbf{z}} f_u(\mathbf{z}_u) \cdot \boldsymbol{\delta}$ to
zero, which geometrically enforces $\nabla_{\mathbf{z}}
f_u(\mathbf{z}_u) \perp \boldsymbol{\delta}$. This result has a
profound consequence: the classifier's decision boundary becomes
strictly orthogonal to the subspace of nuisance factors (semantics
and generator fingerprints). Invariance is achieved because the
gradient along any nuisance direction $\boldsymbol{\delta}$ is
suppressed to zero---the model classifies based solely on universal
forgery evidence $\mathbf{z}_u$, discarding the ``who'' and the
``what'' to focus on the ``whether.'' A complete derivation with
detailed geometric interpretation is provided in Appendix~\ref{appendix:purification}.

\subsection{Manifold Alignment}
\label{sec:alignment}
Generative models induce heterogeneous feature distributions leading to domain gaps that hinder detector generalization. 
To reduce geometric disparity among forgery classes, we employ manifold alignment that regularizes the latent space of purified universal features. We maintain a global centroid $\mathbf{c}_{global}$ representing the aggregate fake class and local centroids $\{\mathbf{c}_k\}_{k=1}^K$ for each specific generator family $k$. The alignment loss contracts local centroids towards the global prototype:
\begin{equation}
\mathcal{L}_{\text{align}} = \frac{1}{K} \sum_{k=1}^K \| \mathbf{c}_k - \mathbf{c}_{global} \|_2^2,
\label{eq:manifold_alignment}
\end{equation}
thereby promoting a compact, unified manifold and enhancing inter-domain robustness.

\subsection{Optimization Objective}
The complete ODP-Net is trained end-to-end with a composite objective that balances classification accuracy, feature sufficiency, robustness, and manifold compactness:
\begin{equation}
    \begin{aligned}
\mathcal{L}_{\text{total}} = & \mathcal{L}_{\text{cls}} + \alpha (\mathcal{L}_{\text{suff}}^{auth} + \mathcal{L}_{\text{suff}}^{gen}) \\ 
+ & \beta \mathcal{L}_{\text{puri}} + \gamma \mathcal{L}_{\text{align}} + \lambda \mathcal{L}_{\text{sparse}},
    \end{aligned}
    \label{eq:total_loss}
\end{equation}
where $\mathcal{L}_{\text{cls}}$ denotes the standard cross-entropy losses for real/fake detection and generator ID classification. The complete training procedure, including the detailed data flow and optimization steps, is summarized in Algorithm~\ref{alg:main_flow}.

\begin{algorithm}[ht]
\caption{ODP-Net Training and Evaluation}
\label{alg:main_flow}
\begin{algorithmic}[1]
\REQUIRE Training set $\mathcal{D}_{\text{train}}$, Unseen test set $\mathcal{D}_{\text{test}}$, Pretrained encoder $E$, Gating network $G$, Classifiers $f_u, f_s$
\STATE \textbf{// --- Phase 1: Training ---}
\FOR{each minibatch $\mathcal{B} \subset \mathcal{D}_{\text{train}}$}
  \STATE \textbf{Decomposition} (Sec.~\ref{sec:method_decomposition})
  \STATE $\mathbf{z} \leftarrow E(\mathbf{x}),\ \forall \mathbf{x} \in \mathcal{B}$
  \STATE $\mathbf{M}_u, \mathbf{M}_s \leftarrow \text{STE}(G(\mathbf{z}))$ \hfill $\triangleright$ Eq.~\ref{eq:ste_mask}
  \STATE $\mathbf{z}_u \leftarrow \mathbf{z} \odot \mathbf{M}_u$
  \STATE $\mathbf{z}_s \leftarrow \mathbf{z} \odot (\mathbf{1} - \mathbf{M}_u) \odot \mathbf{M}_s$
  \STATE $\mathbf{z}_n \leftarrow \mathbf{z} \odot (\mathbf{1} - \mathbf{M}_u) \odot (\mathbf{1} - \mathbf{M}_s)$
  \STATE \textbf{Classification}
  \STATE $p_{\text{fake}} \leftarrow \text{Softmax}(f_u(\mathbf{z}_u))$, \quad $p_{\text{gen}} \leftarrow \text{Softmax}(f_s(\mathbf{z}_s))$
  \STATE $\mathcal{L}_{\text{cls}} \leftarrow \mathcal{L}_{\text{CE}}(p_{\text{fake}}, y) + \mathcal{L}_{\text{CE}}(p_{\text{gen}}, k)$
  \STATE \textbf{Information Sufficiency} (Eq.~\ref{eq:sufficiency})
  \STATE $\mathcal{L}_{\text{suff}} \leftarrow D_{\text{KL}}(f_u(\mathbf{z}_u) \| \text{sg}[f_u(\mathbf{z})]) + D_{\text{KL}}(f_s(\mathbf{z}_s) \| \text{sg}[f_s(\mathbf{z})])$
  \STATE \textbf{Purification} (Sec.~\ref{sec:purification})
  \STATE Sample donor $\mathbf{z}' \neq \mathbf{z}$ from $\mathcal{B}$
  \STATE $\mathbf{z}'_{\text{nuis}} \leftarrow \mathbf{z}' \odot (\mathbf{1} - \mathbf{M}'_u)$ \hfill $\triangleright$ Eq.~\ref{eq:donor_nuisance}
  \STATE $\mathbf{z}_{\text{hybrid}} \leftarrow \mathbf{z}_u + \mathbf{z}'_{\text{nuis}}$ \hfill $\triangleright$ Eq.~\ref{eq:hybrid_feature}
  \STATE $\mathcal{L}_{\text{puri}} \leftarrow \| f_u(\mathbf{z}_u) - f_u(\mathbf{z}_{\text{hybrid}}) \|_2^2$
  \STATE \textbf{Manifold Alignment} (Sec.~\ref{sec:alignment})
  \STATE Update per-generator centroids $\{\mathbf{c}_k\}$ and global centroid $\mathbf{c}_{\text{global}}$
  \STATE $\mathcal{L}_{\text{align}} \leftarrow \frac{1}{K} \sum_{k=1}^K \| \mathbf{c}_k - \mathbf{c}_{\text{global}} \|_2^2$
  \STATE \textbf{Optimization}
  \STATE $\mathcal{L}_{\text{total}} \leftarrow \mathcal{L}_{\text{cls}} + \alpha \mathcal{L}_{\text{suff}} + \beta \mathcal{L}_{\text{puri}} + \gamma \mathcal{L}_{\text{align}} + \lambda \mathcal{L}_{\text{sparse}}$
  \STATE Update $\theta$ via $\nabla_\theta \mathcal{L}_{\text{total}}$
\ENDFOR

\STATE \textbf{// --- Phase 2: Evaluation ---}
\FOR{each sample $\mathbf{x} \in \mathcal{D}_{\text{test}}$}
  \STATE $\mathbf{z} \leftarrow E(\mathbf{x})$
  \STATE $\mathbf{M}_u \leftarrow \text{STE}(G(\mathbf{z}))$ \hfill $\triangleright$ Only universal mask needed
  \STATE $\mathbf{z}_u \leftarrow \mathbf{z} \odot \mathbf{M}_u$
  \STATE $\hat{y} \leftarrow \arg\max\ f_u(\mathbf{z}_u)$ \hfill $\triangleright$ Real/Fake prediction
\ENDFOR
\ENSURE Predictions $\{\hat{y}\}$ and confidence scores $\{p_{\text{fake}}\}$ for all test samples
\end{algorithmic}
\end{algorithm}

\section{Experiments}

\begin{table*}[ht!]
\caption{Quantitative comparison of in-domain detection performance. We evaluate methods across four categories of generative models: GAN-based Noise-to-Image Generation (GAN-N2IG), Diffusion Model for Personalized Generation (DM-PG), Diffusion Model for Text-to-Image Generation (DM-T2IG), and Deepfakes. The evaluation metrics include Balanced Accuracy (bAcc, $\uparrow$) and Negative Log Likelihood (NLL, $\downarrow$). The best and second-best results are highlighted in \textbf{bold} and \underline{underlined}, respectively.}
\label{tab:in-domain}
\centering
\resizebox{2.0\columnwidth}{!}{%
\begin{tabular}{@{}c|cccc|cccc|cccc|cccc|cc@{}}
\toprule
\multirow{3}{*}{Method} & \multicolumn{4}{c|}{GAN-N2IG}                                    & \multicolumn{4}{c|}{DM-PG}                                           & \multicolumn{4}{c|}{DM-T2IG}                                          & \multicolumn{4}{c|}{Deepfakes}                                    & \multirow{3}{*}{mbAcc↑} & \multirow{3}{*}{mNLL↓} \\ \cmidrule(lr){2-17}
                                 & \multicolumn{2}{c}{ProGAN} & \multicolumn{2}{c|}{R3GAN} & \multicolumn{2}{c}{BLIP} & \multicolumn{2}{c|}{Infinite-ID} & \multicolumn{2}{c}{DALLE-3} & \multicolumn{2}{c|}{FLUX1-dev} & \multicolumn{2}{c}{BlendFace} & \multicolumn{2}{c|}{E4S} &                                  &                                 \\
                                 & bAcc↑   & NLL↓    & bAcc↑   & NLL↓    & bAcc↑  & NLL↓   & bAcc↑      & NLL↓       & bAcc↑    & NLL↓    & bAcc↑     & NLL↓      & bAcc↑     & NLL↓     & bAcc↑  & NLL↓   &                                  &                                 \\ \midrule
AIDE                             & 0.8249           & 0.3933           & 0.7589           & 0.5144           & 0.7761          & 0.4673          & 0.8089              & 0.4044              & 0.7259            & 0.5296           & 0.7828             & 0.4490             & 0.9606             & 0.1047            & 0.7706          & 0.4677          & 0.8011                           & 0.4163                          \\
FreqNet                          & 0.9837           & 0.0574           & 0.8406           & 0.3943           & 0.8489          & 0.3931          & 0.8317              & 0.4076              & 0.8114            & 0.5119           & 0.8106             & 0.4567             & \textbf{0.9961}             & \textbf{0.0156}            & 0.8239          & 0.4536          & 0.8684                           & 0.3363                          \\
RINE                             & 0.9948           & {\ul 0.0146}     & 0.9561           & 0.1188           & 0.9628          & 0.1158          & 0.9350              & 0.1627              & 0.9444            & 0.1653           & 0.9583             & 0.1261             & 0.9617             & 0.1338            & 0.9561          & 0.1268          & 0.9587                           & 0.1205                          \\
NRR                              & {\ul 0.9963}     & \textbf{0.0133}  & 0.9433           & 0.1503           & 0.9500          & 0.1357          & 0.9444              & 0.1606              & 0.9232            & 0.2254           & 0.9633             & 0.1132             & 0.9878             & 0.0291            & 0.9572          & 0.1029          & 0.9582                           & 0.1163                          \\
VIB                              & 0.9826           & 0.0549           & 0.9389           & 0.2015           & 0.9872          & 0.0332          & 0.9878              & 0.0332              & 0.9875            & {\ul 0.0378}     & 0.9868             & 0.0412             & 0.8328             & 0.4027            & 0.9883          & 0.0340          & 0.9615                           & 0.1048                          \\
FatFormer                        & 0.9951           & 0.0323           & \textbf{0.995}   & {\ul 0.0286}     & 0.9933          & {\ul 0.0311}    & {\ul 0.9933}        & {\ul 0.0198}        & {\ul 0.9913}      & 0.0443           & 0.9856             & \textbf{0.0268}    & 0.9522             & 0.1337            & {\ul 0.9914}    & {\ul 0.0247}    & {\ul 0.9872}                     & {\ul 0.0427}                    \\
SAFE                             & \textbf{0.9981}  & 0.0511           & 0.9889           & 0.0995           & \textbf{0.9967} & 0.0638          & {\ul 0.9933}        & 0.0892              & 0.5506            & 0.8873           & \textbf{0.9933}    & 0.0754             & 0.7594             & 0.4882            & 0.9594          & 0.1427          & 0.9050                           & 0.2372                          \\ \midrule
Ours                             & 0.9952           & 0.0201           & {\ul 0.9933}     & \textbf{0.0185}  & {\ul 0.9939}    & \textbf{0.0187} & \textbf{0.9967}     & \textbf{0.0093}     & \textbf{0.9925}   & \textbf{0.0197}  & {\ul 0.9878}       & {\ul 0.0392}       & {\ul 0.9922}       & {\ul 0.0213}      & \textbf{0.9917} & \textbf{0.0233} & \textbf{0.9929}                  & \textbf{0.0213}                 \\ \bottomrule
\end{tabular}%
}
\end{table*}

\begin{table*}[ht]
\caption{Quantitative evaluation of cross-generator generalization in terms of Balanced Accuracy (bAcc, $\uparrow$). We assess the model's robustness against unseen generative models across diverse categories: GAN-based (GAN-N2IG), Personalized Diffusion (DM-PG), Text-to-Image Diffusion (DM-T2IG), Face Swapping (Deepfakes), and complex real-world imagery gathered from online sources (In-the-Wild). The best and second-best results are highlighted in \textbf{bold} and \underline{underlined}, respectively.}
\label{tab:cross_domain_bacc}
\centering
\resizebox{2\columnwidth}{!}{%
\begin{tabular}{@{}c|cc|cc|cc|cc|cc|c@{}}
\toprule
\multirow{2}{*}{Method} & \multicolumn{2}{c|}{GAN-N2IG}    & \multicolumn{2}{c|}{DM-PG}        & \multicolumn{2}{c|}{DM-T2IG}      & \multicolumn{2}{c|}{Deepfakes}    & \multicolumn{2}{c|}{In-the-Wild} & \multirow{2}{*}{Avg.} \\ \cmidrule(lr){2-11}
                        & StyleSwim       & WFIR           & InstantID       & IP-Adapter      & Midjourney      & SD3             & FaceSwap        & InSwap          & CommunityAI    & SocialRF        &                       \\ \midrule
RINE                    & 0.4981          & 0.5065         & 0.7889          & 0.8822          & 0.9605          & 0.9448          & 0.9784          & 0.9801          & 0.672          & 0.7162          & 0.79277               \\
NRR                     & 0.564           & 0.501          & 0.8749          & 0.7851          & 0.9242          & 0.9256          & 0.9743          & 0.9786          & 0.6282         & 0.7172          & 0.78731               \\
FreqNet                 & 0.629           & 0.5375         & 0.6293          & 0.8173          & 0.818           & 0.73            & 0.8969          & 0.9052          & 0.583          & 0.6185          & 0.71647               \\
AIDE                    & 0.7364          & 0.7495         & 0.7828          & 0.7754          & 0.7318          & 0.7791          & 0.7818          & 0.787           & 0.4508         & 0.5402          & 0.71148               \\
FatFormer               & 0.7489          & 0.5335         & 0.9316          & 0.8903          & 0.9813          & 0.9854          & {\ul 0.987}     & {\ul 0.9901}    & {\ul 0.745}    & \textbf{0.8258} & 0.86189               \\
VIB                     & 0.8092          & {\ul 0.7935}   & 0.9311          & 0.9552          & {\ul 0.9842}    & {\ul 0.9886}    & 0.9855          & 0.9872          & 0.7046         & 0.789           & {\ul 0.89281}         \\
SAFE                    & {\ul 0.8753}    & 0.496          & \textbf{0.9418} & \textbf{0.9651} & 0.9535          & 0.9147          & 0.9728          & 0.958           & 0.576          & 0.5687          & 0.82219               \\ \midrule
Ours                    & \textbf{0.9322} & \textbf{0.954} & {\ul 0.9394}    & {\ul 0.957}     & \textbf{0.9845} & \textbf{0.9937} & \textbf{0.9941} & \textbf{0.9946} & \textbf{0.749} & {\ul 0.7952}    & \textbf{0.92937}      \\ \bottomrule
\end{tabular}%
}
\end{table*}

\begin{table*}[ht]
\caption{Quantitative evaluation of probability calibration in terms of Negative Log Likelihood (NLL, $\downarrow$). We evaluate the model's predictive reliability when encountering unseen generative models across diverse categories: GAN-based (GAN-N2IG), Personalized Diffusion (DM-PG), Text-to-Image Diffusion (DM-T2IG), Face Swapping (Deepfakes), and complex real-world imagery gathered from online sources (In-the-Wild). The best and second-best results are highlighted in \textbf{bold} and \underline{underlined}, respectively.}
\label{tab:cross_domain_nll}
\centering
\resizebox{2\columnwidth}{!}{%
\begin{tabular}{@{}c|cc|cc|cccc|cc|c@{}}
\toprule
\multirow{2}{*}{Method} & \multicolumn{2}{c|}{GAN-N2IG}    & \multicolumn{2}{c|}{DM-PG}        & \multicolumn{2}{c|}{DM-T2IG}                        & \multicolumn{2}{c|}{Deepfakes}    & \multicolumn{2}{c|}{In-the-Wild}  & \multirow{2}{*}{Avg.} \\ \cmidrule(lr){2-11}
                        & StyleSwim      & WFIR            & InstantID       & IP-Adapter      & Midjourney      & \multicolumn{1}{c|}{SD3}          & FaceSwap        & InSwap          & CommunityAI     & SocialRF        &                       \\ \midrule
RINE                    & 1.5601         & 1.5645          & 0.4145          & 0.2441          & 0.0956          & \multicolumn{1}{c|}{0.1405}       & 0.0573          & 0.0554          & {\ul 0.7943}    & 0.7197          & 0.5646                \\
NRR                     & 2.4913         & 4.1828          & 0.3229          & 0.5523          & 0.2063          & \multicolumn{1}{c|}{0.2519}       & 0.0698          & 0.0544          & 1.4334          & 0.9105          & 1.04756               \\
FreqNet                 & 1.1532         & 1.6916          & 1.2129          & 0.4219          & 0.4739          & \multicolumn{1}{c|}{0.7352}       & 0.268           & 0.2407          & 2.0344          & 1.1194          & 0.93512               \\
AIDE                    & 0.5584         & {\ul 0.5857}    & 0.4929          & 0.494           & 0.5696          & \multicolumn{1}{c|}{0.4799}       & 0.4709          & 0.4525          & \textbf{0.7879} & 0.8135          & 0.57053               \\
FatFormer               & 2.3309         & 7.9292          & 0.3833          & 0.7353          & 0.0965          & \multicolumn{1}{c|}{0.0878}       & {\ul 0.0191}    & {\ul 0.0202}    & 2.7217          & 1.5511          & 1.58751               \\
VIB                     & 0.6153         & 0.7179          & 0.3132          & {\ul 0.1364}    & {\ul 0.0439}    & \multicolumn{1}{c|}{{\ul 0.0331}} & 0.0382          & 0.0385          & 0.8423          & \textbf{0.6278} & {\ul 0.34066}         \\
SAFE                    & {\ul 0.3662}   & 1.5392          & \textbf{0.1744} & \textbf{0.0901} & 0.1519          & \multicolumn{1}{c|}{0.2702}       & 0.1014          & 0.1319          & 0.9962          & 1.0358          & 0.48573               \\ \midrule
Ours                    & \textbf{0.307} & \textbf{0.1998} & {\ul 0.2693}    & 0.1775          & \textbf{0.0427} & \textbf{0.0169}                   & \textbf{0.0162} & \textbf{0.0157} & 0.9197          & {\ul 0.672}     & \textbf{0.26368}      \\ \bottomrule
\end{tabular}%
}
\end{table*}

\begin{figure*}[t!]
\centering
\includegraphics[width=5.5in]{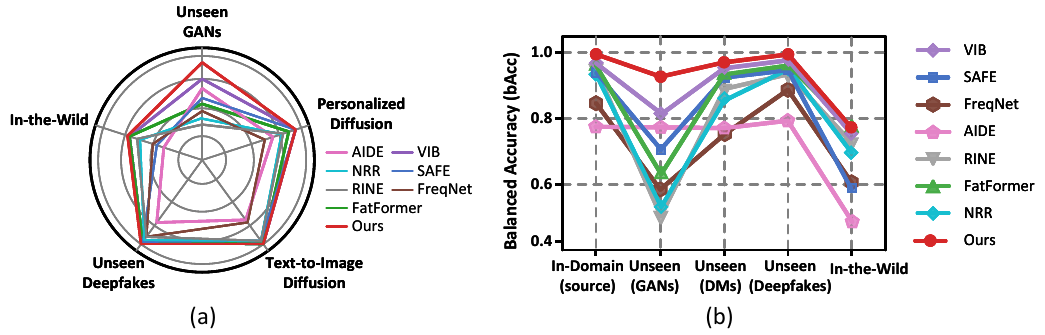}
\caption{Holistic evaluation of generalized detection capabilities. (a) Radar chart comparing the detection sensitivity across five distinct generative categories. ODP-Net (Red) maintains a comprehensive, isotropic performance envelope, whereas baselines exhibit severe collapses on specific families. (b) Performance trajectory across distribution shifts. While other methods experience pronounced declines on unseen GANs, ODP-Net remains stable.}
\centering
\label{fig:generalized}
\end{figure*}

\subsection{Experimental Settings}
\textbf{Datasets and Protocols.}
We conduct a comprehensive evaluation to assess cross-generator generalization and real-world robustness using the AIGIBench dataset~\cite{li2025artificial}. 
The dataset is partitioned into disjoint source and target domains to simulate the realistic scenario where detectors must face unknown synthesis methods. 
The Source Domain (Training) comprises eight diverse architectures: ProGAN~\cite{karras2017progressive}, R3GAN~\cite{huang2024gan}, BLIP~\cite{li2022blip}, Infinite-ID~\cite{wu2024infinite}, DALLE-3~\cite{betker2023improving}, FLUX1-dev~\cite{batifol2025flux}, BlendFace~\cite{shiohara2023blendface}, and E4S~\cite{li2023e4s}. 
The Target Domain (Unseen Evaluation) consists of eight novel generators: StyleSwim~\cite{zhang2022styleswin}, WFIR\footnote{https://www.whichfaceisreal.com/}, InstantID~\cite{wang2024instantid}, IP-Adapter~\cite{ye2023ip}, Midjourney\footnote{https://www.midjourney.com/home}, SD3~\cite{esser2024scaling}, FaceSwap\footnote{https://github.com/deepfakes/faceswap}, and InSwap\footnote{https://github.com/haofanwang/inswapper}. 
Additionally, we evaluate on the In-the-Wild subset (CommunityAI~\cite{li2025artificial} and SocialRF~\cite{li2025artificial}) to assess real-world robustness.
We adopt a diverse multi-source training set because ODP-Net's explicit generator classification objective ($\mathcal{L}_{\text{suff}}^{gen}$) and decomposition mechanism require meaningful cross-generator variance to learn distinct fingerprint and trace subspaces. All baselines are retrained under this same protocol for fair comparison.

\textbf{Evaluation Metrics.}
We employ Balanced Accuracy (bAcc) to measure classification performance and Negative Log Likelihood (NLL) to evaluate probability calibration. bAcc is computed as $\text{bAcc} = \frac{1}{2}(\frac{TP}{TP+FN} + \frac{TN}{TN+FP})$, while NLL is defined as $\text{NLL} = -\frac{1}{N}\sum_{i=1}^{N}[y_i\log(p_i) + (1-y_i)\log(1-p_i)]$, where $y_i$ denotes ground truth and $p_i$ represents predicted probability. 
We favor bAcc over standard accuracy for its robustness in class-imbalanced evaluation scenarios. By averaging per-class recall, bAcc prevents a dominant class from inflating scores. NLL is preferred over rank-based metrics (e.g., AUC) because it penalizes overconfident incorrect predictions. This property is critical for forensic deployment where miscalibrated confidence can be dangerously misleading. A higher bAcc value indicates better classification performance, whereas a lower NLL value reflects more accurate probability calibration.

\textbf{Baselines.} We compare against seven state-of-the-art methods: AIDE~\cite{yan2024sanity}, FreqNet~\cite{tan2024frequency}, RINE~\cite{koutlis2024leveraging}, NRR~\cite{tan2024rethinking}, VIB~\cite{zhang2025towards}, FatFormer~\cite{liu2024forgery}, and SAFE~\cite{li2025improving}. All baselines are retrained on the identical source domain for fair comparison.

\textbf{Implementation.} ODP-Net adopts a pre-trained CLIP ViT-L/14 backbone and trains for 100 epochs using AdamW with cosine annealing ($1\times10^{-6}$ minimum). We apply differential learning rates: $1\times10^{-5}$ for the backbone and $1\times10^{-4}$ for task-specific layers, with $1\times10^{-4}$ weight decay. All loss coefficients ($\alpha, \beta, \gamma, \lambda$) are uniformly set to $1.0$ to demonstrate robustness without extensive tuning.

\subsection{In-Domain Detection Performance}

We first assess whether disentanglement compromises in-domain accuracy. As shown in Table~\ref{tab:in-domain}, ODP-Net avoids the typical trade-off between robustness and precision, achieving competitive results across all source categories while significantly improving probability calibration (lowest NLL in 5/8 subsets).
Baselines reveal distinct biases: spectral methods (e.g., RINE) excel on GANs but falter on diffusion models, while semantic approaches (e.g., SAFE) struggle with text-to-image synthesis (e.g., 0.5506 bAcc on DALLE-3). ODP-Net, however, maintains a consistent performance envelope. The Information Sufficiency constraint ($\mathcal{L}_{\text{suff}}$) ensures that while nuisance factors are filtered, the universal subspace $\mathbf{z}_u$ retains critical discriminative information.

\subsection{Cross-Domain Generalization}

The critical evaluation of any forensic detector lies in its ability to generalize to unseen generative mechanisms. Table~\ref{tab:cross_domain_bacc} (Accuracy) and Table~\ref{tab:cross_domain_nll} (Calibration) present the performance on novel architectures and real-world data, highlighting the structural advantages of ODP-Net.

\textbf{Robustness to Novel Architectures.}
Existing detectors exhibit distinct vulnerability patterns driven by feature entanglement. Frequency-based methods (e.g., RINE, FreqNet) rely heavily on spectral artifacts; consequently, they collapse to near-random guessing (approximately $50\%$-$54\%$ bAcc) on WFIR, where modern GANs have successfully suppressed high-frequency fingerprints. Surprisingly, semantic-aware foundation models like FatFormer also fail on this benchmark ($53.35\%$ bAcc) and suffer from extreme calibration errors (Avg NLL $>1.5$). This reveals a critical insight: without structural disentanglement, even powerful vision-language models overfit to source-specific semantics, failing to generalize when the underlying generative mechanism shifts drastically (e.g., from ProGAN to StyleGAN).

ODP-Net successfully breaks this deadlock. By disentangling universal traces from specific fingerprints, our method is the only one to maintain robustness across the spectrum: it recovers high accuracy on the challenging WFIR dataset ($95.40\%$) where others fail, while simultaneously achieving state-of-the-art performance on novel diffusion architectures like SD3 ($99.37\%$).

\textbf{Resilience in Wild Scenarios.}
Real-world scenarios introduce complex, non-linear degradations that scramble fragile frequency fingerprints. This is evident in the performance of RINE and NRR on the CommunityAI dataset, where their accuracy drops to $67.20\%$ and $62.82\%$, respectively. In contrast, ODP-Net achieves state-of-the-art accuracy ($74.90\%$) and, more importantly, maintains calibration stability with an NLL of $0.9197$, whereas FatFormer's NLL spikes to $2.7217$. This resilience is visually corroborated by the radar chart in Figure~\ref{fig:generalized}(a), where ODP-Net forms the largest and most isotropic performance pentagon. Furthermore, Figure~\ref{fig:generalized}(b) illustrates that while baseline performance trajectories decay rapidly as the distribution shift widens (from In-Domain to Unseen GANs), ODP-Net exhibits a flat, stable trajectory. This empirically validates that our counterfactual purification creates a generator-agnostic feature space.

\begin{table*}[ht]
\caption{Component-wise ablation study validating the effectiveness of ODP-Net. We incrementally evaluate the contributions of Instance-aware Orthogonal Decomposition (OD), Information Sufficiency (InfoSuff), Counterfactual Purification (Puri), and Manifold Alignment (Align). The first row represents the baseline CLIP ViT-L/14 backbone trained with standard cross-entropy loss. The results demonstrate that while OD establishes a strong structural baseline, the synergy of all modules is essential for achieving optimal generalization and probability calibration (lowest NLL), particularly in the challenging In-the-Wild scenario.}
\label{tab:component-ablation}
\centering
\resizebox{2\columnwidth}{!}{%
\begin{tabular}{@{}cccc|cc|cc|cc|cc|cc@{}}
\toprule
\multicolumn{4}{c|}{\textbf{Component Composition}} & \multicolumn{2}{c|}{\textbf{GAN-N2IG}} & \multicolumn{2}{c|}{\textbf{DM-PG}} & \multicolumn{2}{c|}{\textbf{DM-T2IG}} & \multicolumn{2}{c|}{\textbf{Deepfakes}} & \multicolumn{2}{c}{\textbf{In-the-Wild}} \\ \midrule
OD       & InfoSuff       & Puri       & Align      & Avg bAcc↑          & Avg NLL↓          & Avg bAcc↑        & Avg NLL↓         & Avg bAcc↑         & Avg NLL↓          & Avg bAcc↑          & Avg NLL↓           & Avg bAcc↑           & Avg NLL↓           \\ \midrule
         &                &            &            & 0.8582             & 0.6177            & 0.8629           & 0.6094           & 0.9001            & 0.0662            & 0.9049             & 0.0255             & 0.7026              & 1.9260             \\
\checkmark        &                &            &            & \textbf{0.9509}    & 0.4166            & {\ul 0.9366}     & 0.5192           & 0.9834            & 0.0932            & 0.9899             & 0.0507             & 0.7572              & 1.8819             \\
\checkmark        & \checkmark              &            &            & 0.8774             & 0.3285            & 0.8077           & 0.4027           & 0.9382            & 0.1599            & 0.9861             & 0.0603             & 0.6895              & {\ul 0.8359}       \\
\checkmark        &                & \checkmark          &            & 0.9361             & 0.3201            & 0.9086           & 0.3863           & \textbf{0.9911}   & \textbf{0.0273}   & {\ul 0.9964}       & {\ul 0.0124}       & \textbf{0.7865}     & 0.9920             \\
\checkmark        &                &            & \checkmark          & 0.9308             & 0.6136            & 0.8874           & 0.7709           & 0.9888            & 0.0915            & 0.9955             & 0.0388             & 0.7454              & 2.0072             \\
\checkmark        & \checkmark              & \checkmark          &            & 0.9167             & {\ul 0.3146}      & 0.8991           & {\ul 0.3320}     & 0.9868            & 0.0376            & \textbf{0.9965}    & \textbf{0.0120}    & 0.7685              & 0.9682             \\
\checkmark        & \checkmark              & \checkmark          & \checkmark          & {\ul 0.9431}       & \textbf{0.2534}   & \textbf{0.9482}  & \textbf{0.2234}  & {\ul 0.9891}      & {\ul 0.0298}      & 0.9944             & 0.0160             & {\ul 0.7721}        & \textbf{0.7959}    \\ \bottomrule
\end{tabular}%
}
\end{table*}

\begin{table}[ht]
\centering
\caption{Quantitative disentanglement verification. Silhouette Score ($\uparrow$) and kNN Accuracy ($\uparrow$) for Authenticity (Auth.) and Generator Identity (Gen.) classification across different feature subspaces. $\mathbf{z}_u$ achieves near-perfect Auth.\ Silhouette (0.9975) while collapsing Gen.\ Silhouette (0.2460), confirming successful isolation of universal traces.}
\label{tab:disentanglement_metrics}
\resizebox{0.48\textwidth}{!}{%
\begin{tabular}{@{}l|cc|cc@{}}
\toprule
\multirow{2}{*}{\textbf{Feature}} & \multicolumn{2}{c|}{\textbf{Auth.\ Clustering}} & \multicolumn{2}{c}{\textbf{Gen.\ Clustering}} \\
\cmidrule(lr){2-5}
 & Silhouette $\uparrow$ & kNN $\uparrow$ & Silhouette $\uparrow$ & kNN $\uparrow$ \\ \midrule
Entangled ($\mathbf{z}$)    & 0.8694 & 99.88\% & 0.9658 & 99.95\% \\
Universal ($\mathbf{z}_u$) & \textbf{0.9975} & \textbf{99.90\%} & 0.2460 & 60.46\% \\
Specific ($\mathbf{z}_s$)   & 0.2976 & 69.93\% & \textbf{0.9790} & \textbf{99.84\%} \\
Residual ($\mathbf{z}_n$)   & 0.7207 & 99.87\% & 0.9526 & 99.95\% \\ \bottomrule
\end{tabular}%
}
\end{table}

\subsection{Ablation and Mechanism Analysis}
To understand the source of these gains, we deconstruct the contribution of each module in Table~\ref{tab:component-ablation}. The results describe a clear hierarchy of necessity.

\textbf{Component Contributions.}
The foundation of our performance is the \textbf{Instance-aware Orthogonal Decomposition (OD)}. Introducing OD (Row 2) yields the single largest performance jump, boosting accuracy on GAN-N2IG from the baseline $85.82\%$ to $95.09\%$. This confirms our core hypothesis: explicit separation of variables is a prerequisite for generalization.
However, decomposition alone is insufficient for calibration, as indicated by the high NLL ($1.8819$) on In-the-Wild data. The integration of \textbf{Information Sufficiency (InfoSuff)} (Row 3) acts as a regularizer, drastically improving calibration. \textbf{Counterfactual Purification (Puri)} (Row 4) proves critical for semantic-heavy domains. On the content-rich DM-T2IG dataset, purification boosts accuracy to $99.11\%$, validating that the active injection of counterfactual nuisance features ($\mathbf{z}_{hybrid}$) successfully compels the network to unlearn semantic biases. Finally, \textbf{Manifold Alignment (Align)} (Row 5) provides the geometric glue, ensuring that the isolated traces cluster tightly.

However, it is the synergistic combination of all modules (Row 7) that delivers the optimal global performance. By integrating decomposition, purification, and alignment, ODP-Net achieves the lowest calibration error (Avg NLL 0.7959) on the challenging In-the-Wild benchmark while maintaining top-tier accuracy. This demonstrates that structural disentanglement and geometric regularization are mutually reinforcing, essential for robust generalization.

Notably, the ablation reveals nuanced interaction effects that clarify each component's role. Comparing Row 4 (OD + Puri, without InfoSuff) against Row 7 (full model) exposes a critical dynamic: Purification alone achieves strong accuracy on structured domains (DM-T2IG bAcc 0.9911) but suffers significant NLL degradation on In-the-Wild data (0.9920 vs.\ 0.7959). The missing Information Sufficiency regularizer allows the universal subspace to drift toward low-entropy representations that overfit to purification artifacts, producing overconfident yet miscalibrated predictions. Without the stabilizing anchor of self-distillation, $\mathbf{z}_u$ loses discriminative fidelity even as invariance improves. Conversely, Manifold Alignment applied without Purification (Row 5) yields the worst calibration overall (In-the-Wild NLL 2.0072): geometric contraction of entangled features collapses distinct failure modes into overlapping clusters, amplifying rather than resolving ambiguity. Only the full synergy---decomposition for structural separation, sufficiency for information preservation, purification for semantic invariance, and alignment for manifold compactness---resolves this tension between accuracy and calibration.

\begin{table}[t]
\centering
\caption{Analysis of Design Variants on Unseen Generalization. We evaluate the impact of different decomposition and purification strategies on the average performance across all unseen target datasets. 
Hard Masking enforces stricter subspace exclusivity than Soft Attention, while counterfactual injection provides superior semantic robustness compared to standard noise injection.
The ``Ours'' row in each section corresponds to the same full model (Hard Masking $+$ Counterfactual Purification), achieving the best accuracy--calibration trade-off.}
\label{tab:ablation_variants}
\centering
\resizebox{1.0\columnwidth}{!}{%
\begin{tabular}{@{}c|l|cc@{}}
\toprule
\multirow{2}{*}{\textbf{Module Analysis}} & \multicolumn{1}{c|}{\multirow{2}{*}{\textbf{Design Variant}}} & \multicolumn{2}{c}{\textbf{Average Unseen Performance}} \\ \cmidrule(l){3-4} 
 & \multicolumn{1}{c|}{} & \textbf{bAcc} ($\uparrow$) & \textbf{NLL} ($\downarrow$) \\ \midrule
\multirow{3}{*}{\begin{tabular}[c]{@{}c@{}}Decomposition\\ Strategy\end{tabular}} & Random Masking & 0.6433 & 0.3809 \\
 & Soft Attention (Sigmoid) & 0.8779 & 0.2791 \\
 & \textbf{Hard Masking (Ours)} & \textbf{0.9293} & \textbf{0.2674} \\ \midrule
\multirow{3}{*}{\begin{tabular}[c]{@{}c@{}}Purification\\ Strategy\end{tabular}} & Gaussian Noise & 0.8799 & 0.2785 \\
 & Feature Dropout ($p=0.5$) & 0.8789 & 0.2788 \\
 & \textbf{Counterfactual Purification (Ours)} & \textbf{0.9293} & \textbf{0.2674} \\ \bottomrule
\end{tabular}%
}
\end{table}

\textbf{Quantitative Disentanglement Verification.}
To rigorously validate that $\mathbf{z}_u$, $\mathbf{z}_s$, and $\mathbf{z}_n$ indeed capture distinct factors as intended, we compute Silhouette Score and k-Nearest Neighbor (kNN) Accuracy for two tasks: \textbf{Authenticity} (real vs.\ fake) and \textbf{Generator Identity} (classifying fake samples by source) on frozen features from the trained ODP-Net.

As shown in Table~\ref{tab:disentanglement_metrics}, the universal branch $\mathbf{z}_u$ achieves near-perfect Authenticity Silhouette (0.9975) while its Generator identity clustering collapses (0.2460), confirming it discards ``who'' to focus on ``whether.'' Conversely, $\mathbf{z}_s$ excels at Generator clustering (0.9790) while failing at Authenticity (0.2976), verifying that generator-specific fingerprints are successfully isolated. Notably, $\mathbf{z}_n$ retains high Generator Silhouette (0.9526)---different generators exhibit inherent semantic biases (e.g., Midjourney favors artistic styles, GANs often generate faces)---which precisely motivates the Counterfactual Purification module to actively remove these residual correlations.

\textbf{Design Choices and Sensitivity.}
We further rigorously validate our architectural decisions in Table~\ref{tab:ablation_variants}. A critical finding is the superiority of Hard Masking over Soft Attention ($92.93\%$ vs. $87.79\%$ avg bAcc). Soft gating mechanisms suffer from gradient leakage, where the model inadvertently exploits residual information from the suppressed subspace. Hard masking enforces strict orthogonality, preventing this shortcut learning. Similarly, our counterfactual purification strategy outperforms standard Gaussian noise injection ($92.93\%$ vs. $87.99\%$). While random noise is easily filtered, swapping features creates realistic semantic conflicts that force the model to learn true invariance.

\begin{table}[!t]
\centering
\caption{Comparison with alternative augmentation and perturbation strategies on unseen target domains.}
\label{tab:augmentation_compare}
\centering
\resizebox{0.48\textwidth}{!}{%
\begin{tabular}{@{}l|c|c@{}}
\toprule
\textbf{Method} & \textbf{Avg.\ bAcc} ($\uparrow$) & \textbf{Avg.\ NLL} ($\downarrow$) \\ \midrule
MixUp (Input) & 0.8860 & 0.2720 \\
CutMix (Input) & 0.8955 & 0.3057 \\
Gaussian Noise (Feature) & 0.9009 & 0.3483 \\
Feature Dropout (Feature) & 0.9023 & 0.3309 \\
\textbf{Counterfactual Purification (Ours)} & \textbf{0.9293} & \textbf{0.2636} \\ \bottomrule
\end{tabular}%
}
\end{table}

\begin{table}[!t]
\centering
\caption{Robustness evaluation under common image perturbations on unseen target domains.}
\label{tab:robustness}
\begin{tabular}{@{}ll|c|cc|cc@{}}
\toprule
\multirow{2}{*}{\textbf{Method}} & \multirow{2}{*}{\textbf{Metric}} & \multirow{2}{*}{\textbf{Clean}} & \multicolumn{2}{c|}{\textbf{JPEG}} & \multicolumn{2}{c}{\textbf{Blur}} \\
\cmidrule(lr){4-5} \cmidrule(lr){6-7}
 & & & Q=70 & Q=50 & $\sigma$=2 & $\sigma$=3 \\ \midrule
\multirow{2}{*}{SAFE}
 & bAcc$\uparrow$ & 0.8222 & 0.8038 & 0.7941 & 0.7823 & 0.7363 \\
 & NLL$\downarrow$ & 0.4857 & 1.1065 & 1.0991 & 1.1555 & \textbf{1.1583} \\ \midrule
\multirow{2}{*}{\textbf{ODP-Net}}
 & bAcc$\uparrow$ & \textbf{0.9293} & \textbf{0.9029} & \textbf{0.8931} & \textbf{0.8810} & \textbf{0.8338} \\
 & NLL$\downarrow$ & \textbf{0.2636} & \textbf{0.4700} & \textbf{0.5521} & \textbf{0.6489} & 1.3206 \\ \bottomrule
\end{tabular}
\end{table}

We further compare our counterfactual purification with input-level augmentation methods (MixUp, CutMix) and feature-level perturbations (Gaussian noise, feature dropout) in Table~\ref{tab:augmentation_compare}. Input-level mixing can inadvertently destroy subtle high-frequency artifacts, while random feature noise lacks the structured, directed constraints of our counterfactual design. ODP-Net achieves the highest generalization accuracy, confirming that explicit cross-sample semantic injection provides superior invariance learning.

\textbf{Robustness to Common Perturbations.}
Real-world deployment requires robustness to common image degradations. Table~\ref{tab:robustness} evaluates ODP-Net against SAFE under JPEG compression and Gaussian blurring on unseen target domains. While both methods experience accuracy degradation under strong perturbations, ODP-Net maintains substantially more stable probability calibration under JPEG compression and moderate blur---under JPEG Q=50, SAFE's NLL spikes to 1.0991 while ours remains at 0.5521. This confirms that our orthogonal decomposition isolates robust, low-level forgery traces that are particularly resilient to compression artifacts rather than fragile high-frequency patterns easily destroyed by such perturbations.

\begin{table}[!t]
\centering
\caption{Sensitivity analysis of loss hyperparameters. Each weight $\omega$ is varied while others are held fixed at 1.0. The default configuration ($\alpha=\beta=\gamma=\lambda=1.0$) is highlighted in \colorbox{gray!15}{gray}.}
\label{tab:ablation_hyperparam1}
\begin{tabular}{@{}ll|c|c|>{\columncolor{gray!15}}c|c|c@{}}
\toprule
\multirow{2}{*}{\textbf{Component}} & \multirow{2}{*}{\textbf{Metric}} & \multicolumn{5}{c}{\textbf{Weight Value ($\omega$)}} \\
\cmidrule(lr){3-7}
 & & 0.1 & 0.5 & 1.0 & 2.0 & 5.0 \\ \midrule
\multirow{2}{*}{Sufficiency ($\alpha$)}
 & bAcc$\uparrow$ & .9341 & .9257 & .9293 & .9135 & .9198 \\
 & NLL$\downarrow$ & .2488 & .3118 & .2637 & .3235 & .2368 \\ \midrule
\multirow{2}{*}{Purification ($\beta$)}
 & bAcc$\uparrow$ & .9335 & .9206 & .9293 & .9191 & .9314 \\
 & NLL$\downarrow$ & .2553 & .3325 & .2637 & .3209 & .2350 \\ \midrule
\multirow{2}{*}{Alignment ($\gamma$)}
 & bAcc$\uparrow$ & .9272 & .9287 & .9293 & .9303 & .9292 \\
 & NLL$\downarrow$ & .2643 & .2639 & .2637 & .2634 & .2637 \\ \midrule
\multirow{2}{*}{Sparsity ($\lambda$)}
 & bAcc$\uparrow$ & .9213 & .9237 & .9293 & .9217 & .9237 \\
 & NLL$\downarrow$ & .3040 & .3357 & .2637 & .2477 & .2382 \\
\bottomrule
\end{tabular}
\end{table}

\begin{table}[!t]
\centering
\caption{Ablation of backbone contribution. Average bAcc ($\uparrow$) and NLL ($\downarrow$) across all unseen target domains. ODP-Net on ViT-B/16 outperforms standard fine-tuned ViT-L/14, confirming that gains stem from structural disentanglement rather than backbone capacity alone.}
\label{tab:backbone_ablation}
\centering
\resizebox{0.48\textwidth}{!}{%
\begin{tabular}{@{}l|c|c@{}}
\toprule
\textbf{Configuration} & \textbf{bAcc} ($\uparrow$) & \textbf{NLL} ($\downarrow$) \\ \midrule
CLIP ViT-B/16 + Standard Fine-tuning & 0.8064 & 1.7703 \\
CLIP ViT-B/16 + \textbf{ODP-Net} & \textbf{0.8921} & \textbf{0.2746} \\
CLIP ViT-L/14 + Standard Fine-tuning & 0.8455 & 1.6995 \\
CLIP ViT-L/14 + \textbf{ODP-Net} & \textbf{0.9293} & \textbf{0.2636} \\ \bottomrule
\end{tabular}%
}
\end{table}

Table~\ref{tab:ablation_hyperparam1} further examines loss weight sensitivity by sweeping a single $\omega \in \{0.1, 0.5, 1.0, 2.0, 5.0\}$ while holding all others at $1.0$. A consistent trade-off emerges: across every component, smaller weights ($\omega=0.1$) favor higher bAcc while larger weights ($\omega=5.0$) yield better calibration (lower NLL), reflecting the inherent tension between classification accuracy and probability confidence. Consequently, no single $\omega$ simultaneously optimizes both metrics. We therefore adopt $\alpha=\beta=\gamma=\lambda=1.0$ as the neutral midpoint of this scale range, avoiding selective tuning toward either extreme. Crucially, the performance landscape is near-flat: bAcc deviates by at most $1.6$ percentage points from the default across the full $50\times$ weight range, confirming that ODP-Net does not demand delicate hyperparameter coordination.

\textbf{Backbone Independence.}
To isolate the contribution of our proposed modules from the capacity of the CLIP backbone, we compare ODP-Net with standard fine-tuning on both CLIP ViT-B/16 and ViT-L/14 under the cross-generator protocol.

As shown in Table~\ref{tab:backbone_ablation}, applying ODP-Net to the lightweight ViT-B/16 provides a substantial +8.57\% bAcc boost and reduces NLL from 1.7703 to 0.2746, demonstrating that our method's benefits are agnostic to model scale. Critically, ODP-Net on ViT-B/16 (\textbf{0.8921} bAcc) already outperforms the standard fine-tuned, larger ViT-L/14 baseline (\textbf{0.8455} bAcc). This confirms that the generalization power of ODP-Net stems from its orthogonal decomposition and purification design rather than the inherent capacity of a large vision-language model.

\section{Conclusions}
In this work, we attribute the lack of generalization in existing AI-generated image detectors to the entanglement of universal forgery traces with source-specific nuisances. Leveraging the observation that forgery artifacts exhibit spectral additivity, we proposed ODP-Net to structurally decouple these components via instance-aware decomposition and counterfactual purification. By effectively isolating universal forgery traces from semantic and generator-specific factors, our method achieves state-of-the-art performance and probability calibration on unseen architectures, validating the effectiveness of physically grounded feature disentanglement.


%





\ifCLASSOPTIONcaptionsoff
  \newpage
\fi



%
\bibliographystyle{IEEEtran}
\bibliography{IEEEfull,IEEEabrv,IEEEexample}
%
%
%
%
%
%
%
\vfill









\appendices
\section{Spectral Additivity: Theoretical Basis and Empirical Verification}
\label{appendix:spectral_verification}

\textit{This section provides extended qualitative analysis and detailed protocol descriptions complementing the quantitative verification in the main text Section~\ref{sec:spectral_justification}.}

The core premise of ODP-Net is that generator-specific fingerprints and universal forgery traces can be structurally disentangled in the feature space. In this section, we provide the empirical basis for this hypothesis through a rigorous spectral analysis. We demonstrate that artifacts from different generative architectures occupy approximately disjoint frequency subspaces and exhibit additive behavior. This physical additivity motivates our modeling of the latent space as a superposition of independent components.

\subsection{Verification via Differential Analysis}
We validate the additivity hypothesis by analyzing the difference maps between diverse generator spectra. First, considering the difference between a specific generator and the real domain, we isolate the artifact. For example, with ProGAN, we observe $| S_{\text{real}} - S_{\text{gen}_{\text{GAN}}} | \approx | \Phi - (\Phi + \Delta_{\text{stripe}}) | = | \Delta_{\text{stripe}} |$. This operation recovers the isolated vertical striping pattern.
However, the more significant observation arises when comparing two different generative models, such as ProGAN ($A$) and Stable Diffusion ($B$). The spectral difference is given by:
\begin{equation}
| S_{\text{gen}_A} - S_{\text{gen}_B} | \approx | (\Phi + \Delta_{\text{stripe}}) - (\Phi + \Delta_{\text{spot}}) | = | \Delta_{\text{stripe}} - \Delta_{\text{spot}} |.
\end{equation}
If these artifacts occupied the same frequency subspace and interfered destructively, the resulting pattern would be a complex, unrecognizable noise. Instead, as evidenced in Figure~\ref{fig:fft_analysis}, the difference map displays both the vertical stripes of ProGAN and the spectral spots of Stable Diffusion simultaneously and clearly. This visual evidence suggests a phenomenon of spectral additivity, where distinct fingerprints coexist without interference.

\subsection{Connection to ODP-Net Formulation}

The verified spectral additivity provides the physical motivation for the structural decomposition in ODP-Net. A condensed presentation of this argument, including the quantitative validation via the Additivity Criterion, appears in the main text Section~\ref{sec:spectral_justification}.

\section{Theoretical Analysis of Counterfactual Purification}
\label{appendix:purification}

\textit{This section provides the complete derivation complementing the condensed proof in the main text Section~\ref{sec:purification}.}

In this section, we provide a rigorous mathematical justification for the Counterfactual Purification module proposed in Sec.~\ref{sec:purification}.
We aim to prove that minimizing the purification loss $\mathcal{L}_{\text{puri}}$ is theoretically equivalent to enforcing the orthogonality between the decision boundary of the classifier and the subspace of nuisance factors (semantics and specific fingerprints). This orthogonality guarantees that the extracted universal features $\mathbf{z}_u$ become invariant to domain shifts.

\subsection{Problem Formulation}
Let $f_u: \mathbb{R}^d \to \mathbb{R}^C$ denote the authenticity classifier parameterized by $\theta$, mapping the feature space to class logits. Consider a universal feature vector $\mathbf{z}_u$ and a perturbation vector $\boldsymbol{\delta}$.
In our framework, the perturbation $\boldsymbol{\delta}$ is explicitly constructed from the nuisance components of a donor sample (Eq.~\ref{eq:donor_nuisance} and Eq.~\ref{eq:hybrid_feature}):
\begin{equation}
    \boldsymbol{\delta} = \mathbf{z}'_{\text{nuis}} = \mathbf{z}'_s + \mathbf{z}'_n,
\end{equation}
where $\boldsymbol{\delta}$ represents the direction of semantic variation and generator-specific fingerprints, which are assumed to be independent of the forgery traces.

The purification objective (Eq.~\ref{eq:purification_loss}) imposes a consistency constraint:
\begin{equation}
    \min_{\theta} \mathcal{L}_{\text{puri}} = \min_{\theta} \frac{1}{2} \| f_u(\mathbf{z}_u) - f_u(\mathbf{z}_u + \boldsymbol{\delta}) \|_2^2.
    \label{eq:appendix_loss}
\end{equation}

\subsection{Gradient Orthogonality Derivation}

To analyze the behavior of $f_u$ under the perturbation $\boldsymbol{\delta}$, we assume that the function $f_u$ is differentiable and sufficiently smooth (Lipschitz continuous gradients). We perform a first-order Taylor series expansion of $f_u(\mathbf{z}_u + \boldsymbol{\delta})$ around $\mathbf{z}_u$:
\begin{equation}
    f_u(\mathbf{z}_u + \boldsymbol{\delta}) = f_u(\mathbf{z}_u) + \nabla_{\mathbf{z}} f_u(\mathbf{z}_u)^\top \boldsymbol{\delta} + \mathcal{O}(\|\boldsymbol{\delta}\|^2),
    \label{eq:taylor}
\end{equation}
where $\nabla_{\mathbf{z}} f_u(\mathbf{z}_u) \in \mathbb{R}^{d \times C}$ is the Jacobian matrix representing the sensitivity of the classifier to changes in the input features.

Substituting the approximation from Eq. \ref{eq:taylor} into the objective function Eq. \ref{eq:appendix_loss} and ignoring high-order terms $\mathcal{O}(\|\boldsymbol{\delta}\|^2)$ for small perturbations, the loss simplifies to:
\begin{equation}
    \begin{aligned}
    \mathcal{L}_{\text{puri}} &\approx \frac{1}{2} \left\| f_u(\mathbf{z}_u) - \left( f_u(\mathbf{z}_u) + \nabla_{\mathbf{z}} f_u(\mathbf{z}_u)^\top \boldsymbol{\delta} \right) \right\|_2^2 \\
                       &= \frac{1}{2} \left\| \nabla_{\mathbf{z}} f_u(\mathbf{z}_u)^\top \boldsymbol{\delta} \right\|_2^2.
    \end{aligned}
    \label{eq:simplified_loss}
\end{equation}

The minimization of Eq. \ref{eq:simplified_loss} with respect to the feature encoder parameters implies minimizing the dot product between the gradient of the decision function and the perturbation direction.
From the geometric definition of the dot product $\mathbf{a}^\top \mathbf{b} = \|\mathbf{a}\| \|\mathbf{b}\| \cos \theta$, minimizing the magnitude is equivalent to driving $\cos \theta \to 0$ (assuming non-trivial gradients and perturbations).

Consequently, the optimization objective enforces:
\begin{equation}
    \nabla_{\mathbf{z}} f_u(\mathbf{z}_u) \perp \boldsymbol{\delta}.
    \label{eq:orthogonality}
\end{equation}

\subsection{Geometric Interpretation and Invariance}
The result in Eq. \ref{eq:orthogonality} provides a profound geometric interpretation of our method:

\paragraph{Sensitivity Suppression.} The term $\nabla_{\mathbf{z}} f_u(\mathbf{z}_u)$ represents the direction of maximum information gain for the classifier. The perturbation $\boldsymbol{\delta}$ spans the subspace of semantic content and specific fingerprints.

\paragraph{Orthogonal Decision Boundary.} By enforcing $\nabla_{\mathbf{z}} f_u(\mathbf{z}_u) \perp \boldsymbol{\delta}$, we force the decision boundary of $f_u$ to be locally orthogonal to the manifold of nuisance factors.

\paragraph{Semantic Invariance.} Invariance is mathematically defined as zero gradient along specific directions. Since the gradient along the nuisance direction $\boldsymbol{\delta}$ is suppressed to zero, the classifier's prediction becomes invariant to changes in semantics ($\mathbf{z}_n$) or generator identities ($\mathbf{z}_s$).

\paragraph{Conclusion.} This derivation proves that the Counterfactual Purification module does not merely act as data augmentation. Instead, it serves as a regularization term that explicitly disentangles the decision process from nuisance factors. It ensures that the model classifies images based solely on the universal traces $\mathbf{z}_u$, thereby achieving the generalization goal of ``discarding the \textbf{who} and the \textbf{what} to focus on the \textbf{whether}.''

\end{document}